\pdfoutput=1
\documentclass[letterpaper, 10 pt, conference]{ieeeconf}  

\IEEEoverridecommandlockouts                              

\usepackage{graphics} 
\usepackage{epsfig} 
\usepackage{xcolor}
\usepackage{booktabs}
\usepackage{amssymb}
\usepackage{float}
\usepackage{amsmath}
\usepackage{algpseudocode}
\usepackage{algorithm}
\usepackage{hyperref}

\newcommand{\fig}[1]{Fig.~\ref{fig:#1}}
\newcommand{\tabl}[1]{Table~\ref{table:#1}}
\newcommand{\sect}[1]{Sec.~\ref{sec:#1}}

\newcommand*{\DRAFT}{}

\newcommand{\authorspace}{\qquad}
\DeclareRobustCommand*{\IEEEauthorrefmark}[1]{%
  \raisebox{0pt}[0pt][0pt]{\textsuperscript{\footnotesize #1}}%
}

\title{\LARGE \bf
Learning Actionable Representations from Visual Observations}

\author{
Debidatta Dwibedi \IEEEauthorrefmark{R}\thanks{\IEEEauthorrefmark{R} Google AI Resident (\href{http://g.co/airesidency}{g.co/airesidency})} \authorspace Jonathan Tompson \authorspace Corey Lynch \IEEEauthorrefmark{R}\authorspace Pierre Sermanet
\\[0.3em]
Google Brain
}

\begin{document}

\maketitle

\ifdefined\DRAFT
  \thispagestyle{plain}
  \pagestyle{plain}
\else
  \thispagestyle{empty}
  \pagestyle{empty}
\fi

\begin{abstract}
 In this work we explore a new approach for robots to teach themselves about the world simply by observing it. In particular we investigate the effectiveness of learning task-agnostic representations for continuous control tasks. We extend Time-Contrastive Networks (TCN) that learn from visual observations by embedding multiple frames jointly in the embedding space as opposed to a single frame. We show that by doing so, we are now able to encode both position and velocity attributes significantly more accurately. We test the usefulness of this self-supervised approach in a reinforcement learning setting. We show that the representations learned by agents observing themselves take random actions, or other agents perform tasks successfully, can enable the learning of continuous control policies using algorithms like Proximal Policy Optimization (PPO) using only the learned embeddings as input. We also demonstrate significant improvements on the real-world \textit{Pouring} dataset with a relative error reduction of 39.4\% for motion attributes and 11.1\% for static attributes compared to the single-frame baseline. Video results are available at \href{https://sites.google.com/view/actionablerepresentations/}{https://sites.google.com/view/actionablerepresentations}
\end{abstract}

\section{Introduction}

Supervised learning is a powerful approach for learning useful visual representations for many vision tasks, such as image classification, object detection, and semantic segmentation. As such, many state-of-the-art deep-learning approaches for computer vision incorporate pre-training on a supervised surrogate loss, where ample training data is readily available; most commonly the large-scale ImageNet~\cite{deng2009imagenet} or COCO~\cite{lin2014microsoft} classification datasets.

It would be useful if one could likewise take advantage of supervised pre-training for image observation-based robotic control. However, there are three problems that complicate the application of pre-training in this domain. Firstly, it is not entirely clear what semantic labels or surrogate loss is applicable to the robotic control task at hand. In particular, semantic labeling of a scene is often unrelated to robotic control. The set of invariances required for strong classification performance (such as invariance to object pose, intra-class variation, etc) may be unsuitable for a robotic policy. Furthermore, such latent factors are intrinsically necessary to solve the task. Secondly, constructing a data collection pipeline and collecting labels for every new robotic task is prohibitively expensive. Finally, most robotic policies execute on domains that have little overlap with those of large-scale datasets like ImageNet or COCO; they will for instance have little or no coverage on robotic laboratory environments. Additionally, when a general-purpose robot discovers a new environment with previously unseen objects, it needs to adapt and learn about it. This image domain shift often necessitates additional data collection.
Another motivation for learning solely from observation as opposed to learning with access to states or task rewards is to tap into new learning signals. For example one can learn to predict how things move in the world even for things that are out of its control. And this is not limited to learning from demonstration but also from non-demonstration events or objects in the world.

\begin{figure}[tb]
  \centering
  \includegraphics[width=.45\textwidth]{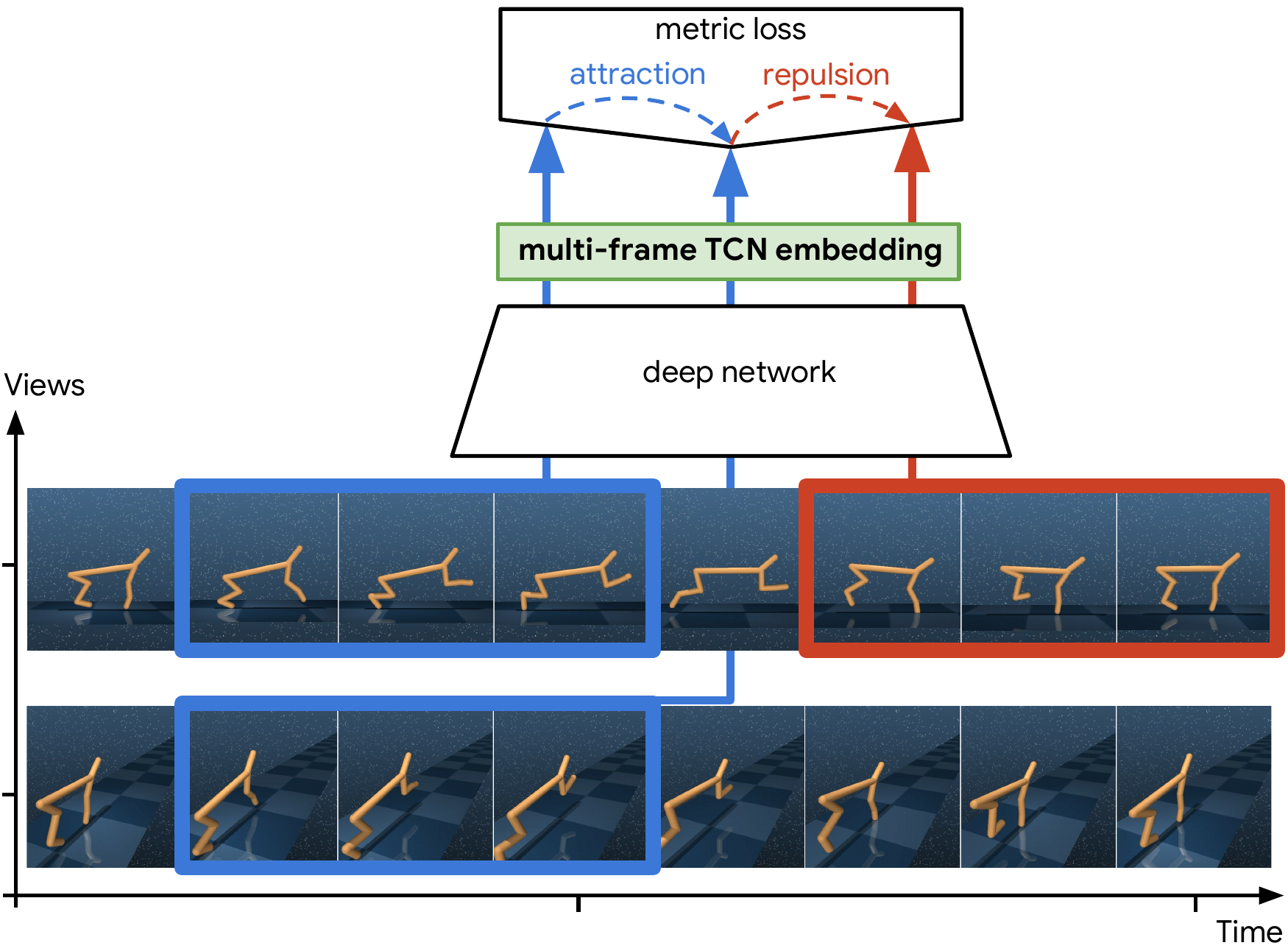}
  \caption{Given two viewpoints of the same event, we sample clips from the video and embed them to produce multi-frame Time-Contrastive Network (mfTCN) embeddings for each view. In order to train the deep network, we consider the clips at the same time from different views to be similar and clips from different time steps to be different.}
  \label{fig:mftcn}
\end{figure}

As an alternative to supervised learning, recent self-supervised approaches, such as Time-Contrastive Networks~\cite{sermanet2017time} (TCN) or Position Velocity Encoders~\cite{jonschkowski2017pves} (PVE), have shown encouraging results when it comes to constructing robust visual representations suitable for Reinforcement Learning (RL) robotics applications, without the need for expensive supervised labels. These methods make use of readily available images directly from the domain of interest. They do so by constructing surrogate metric-learning losses that incorporate a number of structural priors (like temporal consistency, view invariance, etc.) that have been shown to efficiently and compactly encode latent state factors required for policy learning. PVE enabled learning continuous control policies in simulated environments while TCN was shown to help in imitation learning where a robot arm is used to pour liquids into a vessel. However, in the case of TCN, multi-frame state is not encoded, as the embedding is conditioned on a single frame. Likewise, PVE suggest a set of priors that must be tuned for each environment. Additionally, self-supervised techniques have yet to match performance of policies trained directly on true state (rather than image observations). 

However, despite the limitations of recent self-supervised techniques, such approaches learn representations that have a number of desirable properties that we wish to exploit in this work:
1. the embedding can be used to discriminate between different states (including motion attributes) visited in the course of an observation without the need for explicit state labels.
2. the embedding should be robust to changes in viewpoint, thus enabling third-person learning-from-demonstration for which large amounts of readily available training data exists (e.g. YouTube videos).
3. the embedding should be amenable to online adaption in new environments, without the need for additional labels.

In this work we present a variant of TCN, which we call multi-frame Time-Contrastive Network (mfTCN), that encodes temporal latent state (such as velocity, angular velocity, acceleration, etc). We show empirically that our embedding efficiently encodes multi-frame latent state from both real-world and simulation data, by showing that simple linear regressors and classifiers are able to obtain accurate state estimates from a mfTCN embedding. Finally, we show that this demonstration can be used as a basis for PPO\cite{schulman2017proximal} trained policies on simulated robotic control tasks in DeepMind Control Suite~\cite{tassa2018deepmind}, and that mfTCN-based policies reward performance matches those trained from exact state representations.

The contributions of this paper are:
\begin{itemize}
    \item Introducing a multi-frame variant of TCN which works better for both static and motion attributes classification.
    \item Showing RL policies can be learned from pixels using mfTCN while outperforming pixel-based ones learned from scratch or using PVEs.
    \item We also show that the learned policies are competitive with true (proprioceptive) state based policies. Hence, we refer to our representations as \textit{actionable} as they not only encode both static and motion information present in the proprioceptive state but can also be used directly for continuous control tasks.
\end{itemize}

\section{Related Work}

\textbf{Continuous control environments} State-of-the-art performance of Reinforcement Learning (RL) algorithms has improved significantly in recent years and end-to-end, from-pixel policies have shown success on a number of benchmarks including Atari games~\cite{mnih2015human} and continuous control~\cite{lillicrap2015continuous}. Tassa et al.~\cite{tassa2018deepmind} introduced a compelling set of benchmark environments called Control Suite, which we make use of in this work. They are based on tasks initially introduced in the Mujoco environment by Todorov et al.~\cite{todorov2012mujoco}. Another popular benchmark for similar tasks is OpenAI's Gym~\cite{brockman2016openai}. 

\textbf{Learning state representations using priors} Learning useful state representations, in either an unsupervised or semi-supervised setting, has been a long-studied field in robotic control. Scholz et al.~\cite{scholz2014physics} showed that incorporating physics-based priors in the input state  improved performance for model-based RL. Jonschkowski et al.~\cite{jonschkowski2015learning} used similar physics-based \emph{robotic-priors} to learn state representations consistent with the dynamics of the physical task. This framework was extended in \cite{jonschkowski2017pves} to include additional prior terms to capture multi-frame dynamics and where state representations were learned from visual observations only. Similarly to our work, state representations are learned from the observations produced by random agent actions in a simulated environment. Lesort et al.~\cite{lesort2017unsupervised} introduce the reference point prior in their work in addition to the priors introduced in \cite{jonschkowski2017pves}. Concurrent to this work, \cite{pmlr-v87-amiranashvili18a} also demonstrate the importance of representing motion in continuous control tasks. They do so by using the  optical flow predicted by a network as an additional input to the policy. 

\textbf{Self-supervised learning using  images} There have been multiple successes in using unsupervised and semi-supervised pre-training approaches to learn visual representations useful for the task of robotics and reinforcement learning in recent years. Watter et al.~\cite{watter2015embed} learn a locally linear latent space that allows them  use optimal control algorithms to follow trajectories in the embedding space. van Hoof et al.~\cite{van2016stable} use variational auto-encoders to stabilize a reinforcement learning system based on visual and tactile data streams. Finn et al.~\cite{finn2016deep} successfully learn a model that encodes an input image in a low dimensional space. The model is trained to reconstruct the input image. The learned embedding is provided as input along with the true state of the robot. This joint representation enables the robot to preform complicated tasks like rice scooping and looping hooks which were not possible without the visual input. Munk et al.~\cite{munk2016learning} propose an approach to map their input to a useful hidden state using "predictive priors" before training their Actor-Critic model using reinforcement learning. Agrawal et al.~\cite{agrawal2016learning} introduce an interesting framework in which an agent first learns a model of the world by observing the effect of the random actions it takes. This learned representation can then be used to perform tasks that require multi-step decision making. Pinto et al.~\cite{pinto2016curious} show that it is possible to learn good visual representations by taking pre-defined actions with a robot in the real world. As the learned representations get better, this should also enable the robot to perform more complicated tasks. Recent work on imitation learning Yu et al.~\cite{yu2018one} even advocates directly learning policies from pixels directly by performing meta-learning to adapt to different demonstrations.

\textbf{Self-supervised learning on videos} Researchers have also had success in learning useful representation by using videos as input. Sermanet et al.~\cite{sermanet2017time} use time as a supervisory signal to learn the structure present in videos to learn a robust task-agnostic visual representation. Pathak et al.~\cite{pathakICLR18zeroshot} also train a visual classifier that helps an agent identify intermediates states an agent needs to visit to complete a task. This classifier is trained using an self-supervised training objective based on temporal coherency. Finn et al.~\cite{finn2016unsupervised} introduced a dataset of robot-object interactions in the real-world. They learn concise visual representations by predicting the future frames in a video. Babaeizadeh et al.~\cite{babaeizadeh2017stochastic} extend the above work by predicting the future frames in a stochastic manner. Dosovitskiy et al. \cite{dosovitskiy2016learning} show that they can learn to take actions in an environment by predicting future state changes.

\textbf{Auxiliary losses to improve representations} While the above approaches highlight the advantage of using robust representations learned prior to the reinforcement learning, there has been recent work on adding auxiliary tasks to the RL objective to learn models that perform better~\cite{shelhamer2016loss, jaderberg2016reinforcement, mirowski2016learning}. Multi-task learning has also been shown to be helpful in learning more robust internal representations that ultimately lead to improvement in performance~\cite{pinto2017learning,arora2018multi,teh2017distral}. While in this work, we only consider the scenario where we learn task-agnostic representations before learning the control policy, we can adapt our approach to a multi-task setting where the policy and representation learning both improve jointly. 

\textbf{Time-Contrastive Networks (TCNs)} In this work we extend TCNs introduced by Sermanet et al.~\cite{sermanet2017time} which uses time as a supervisory signal. Using TCNs it is possible to discriminate between various states that an agent might encounter while leanring to perform a task. The representations have been shown to be useful for imitation learning tasks. In traditional control systems, the input is the position and velocity of various objects or parts of the agent. Since, we will be using the learned representation to train agents to learn control policies, one desirable property of the embedding would be to encode both position and velocity. In the original TCN, the model was capable of encoding the state of the world and the position of various objects/agent parts. However, it was difficult for TCN to encode motion cues or velocity of objects/agent parts because it was conditioned on only a single frame. In this work, we extend TCNs to the multi-frame setting by  embedding multiple-frames at each time step we expect to learn a representation that encodes both static and motion attributes.

\section{Approach}

Videos have the potential to be a rich source of data for a robot increasing the knowledge about its environment many fold. It is however difficult to learn control policies directly from pixels based only on the rewards that an agent receives. This is where robust visual representations can come in handy. Learning from visual observations seems to be a key ingredient in how humans pick up motor skills. Often times humans transfer skills to each other by performing demonstrations of the task. We also see examples of doctors teleoperating surgical instruments using visual feedback as opposed to having direct access to the position and velocities of the instruments. We envision a future where robots will be able to take advantage of visual feedback in conjunction with input from their other sensors. A good visual representation might also hold the key for transfer of skill from robots to humans via demonstrations in virtual reality. From a continual learning perspective, we want to learn a visual representation that gets updated even when no demonstrations are provided. The agent learns by observing changes in its environment. To that end, we propose a task-agnostic approach to learn such representations from observations that allows us to learn control policies from pixels. We essentially disentangle the representation learning phase from the task-specific control policy learning phase. We present details for both these sections below.

\begin{figure}[tb]
  \centering
  \includegraphics[width=0.5\textwidth]{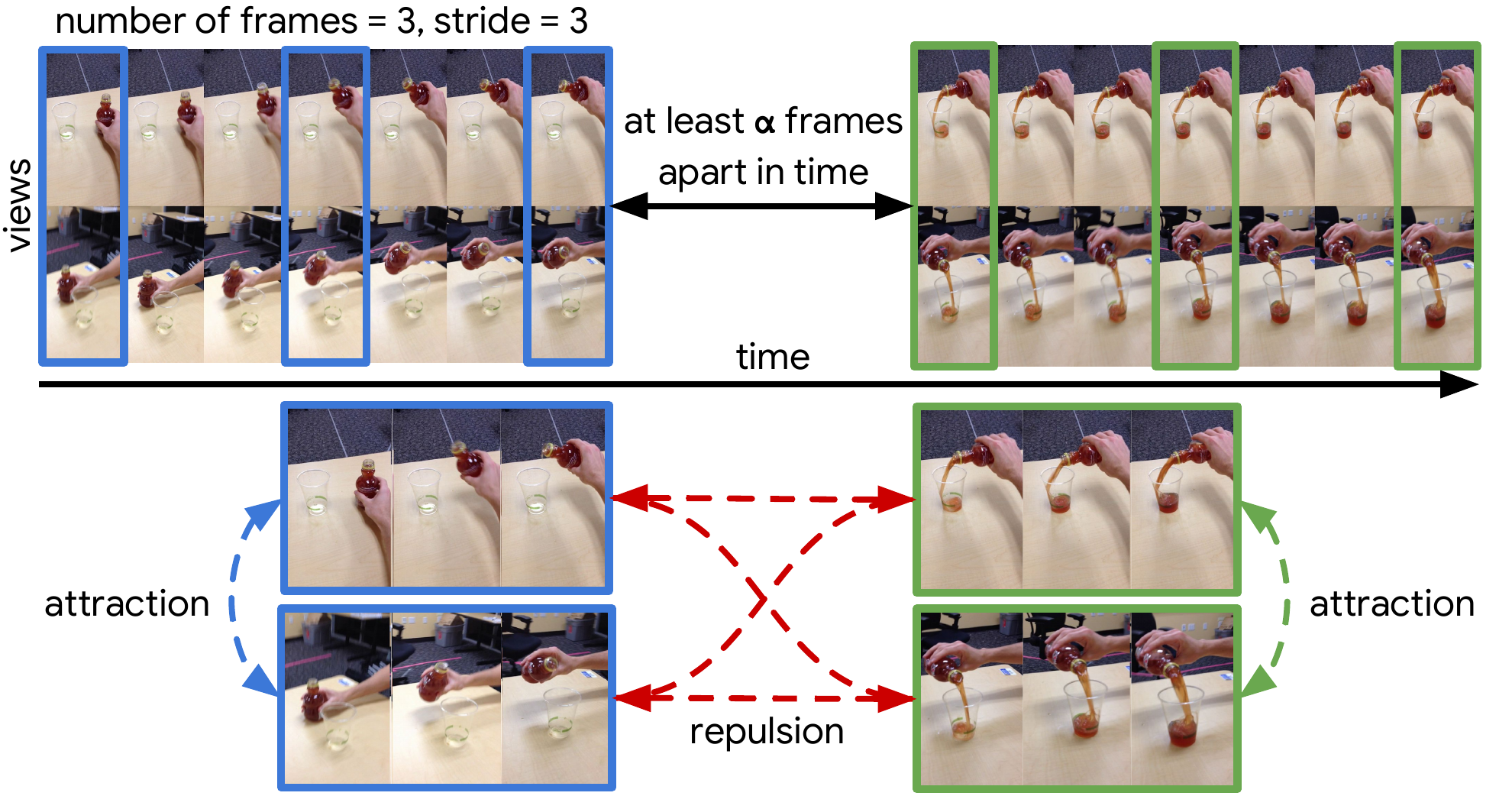}
  \caption{In the above figure, we showcase how we sample a batch for training our model. As shown in the top row, we sample clips from both views simultaneously where each clip consists of 3 frames with a stride of 3 between each frame. We ensure that the clips are not overlapping and have at least $\alpha$ time steps gap between them. With just two samples, we can see the strength of the self-supervisory signal which forces the model to look for differences in similar looking frames while also looking for similarities between different views.}
  \label{fig:sampling}
\end{figure}

\subsection{Learning Representations from Observations}

In this phase, an agent first learns representations in  a task-agnostic manner. The agent can learn from a variety of observations: from passive observations of its environment, from demonstrations by other agents, and from observing itself act in the world. This task-agnostic learning is versatile in the sense that it allows the model to learn from the world and from other agents (i.e. even when it does not have access to true state or even rewards) but also from itself. In this work, we restrict ourselves to the multi-view setting where for each observation we have two synchronized views (the model could be trained with more views as well).

Formally, we are provided with multiple videos as input from which an agent can observe and learn good representations of the world. Each video $v$ has been collected from multiple synchronized viewpoints $v_1, v_2, ... , v_k$. As in ~\cite{sermanet2017time}, we use time as a supervisory signal. In Figure ~\ref{fig:sampling}, we show how we create a training batch from the given videos. We consider clips at time $t$ from all given viewpoints to be similar to each other (they will attract each other in embedding space). Additionally, these clips are also considered dissimilar to any clip that is beyond $\alpha$ steps in time in any view (dissimilar clips will repulse each other in embedding space). To encourage the network to learn representations that encode the above intuition, we learn a metric space with the embeddings produced by a base network. Unlike \cite{sermanet2017time}, instead of embedding a single frame at each timestep we embed multiple frames at each timestep. To do so, we introduce two hyperparameters: number of frames that are embedded together denoted as $n$ and the stride between these frames denoted as $s$. At each time step t, we embed the current frame and the previous $n-1$ frames chosen with a stride of $s$ between the frames. This means at each timestep the network has a fixed lookback window of $(n-1) \times s + 1$ frames.

The motivation for embedding multiple frames is to allow the network to reason not only about the states of objects but also exploit the motion cues present in a scene. As demonstrated in \sect{experiments} and predictably, TCN has difficulty encoding motion since it embeds a single frame at a time. We alleviate this problem in the multi-frame version by embedding multiple frames jointly, which makes it easier to encode motion cues and velocity of objects.

\begin{figure*}[tb]
  \centering
  \includegraphics[width=0.8\textwidth]{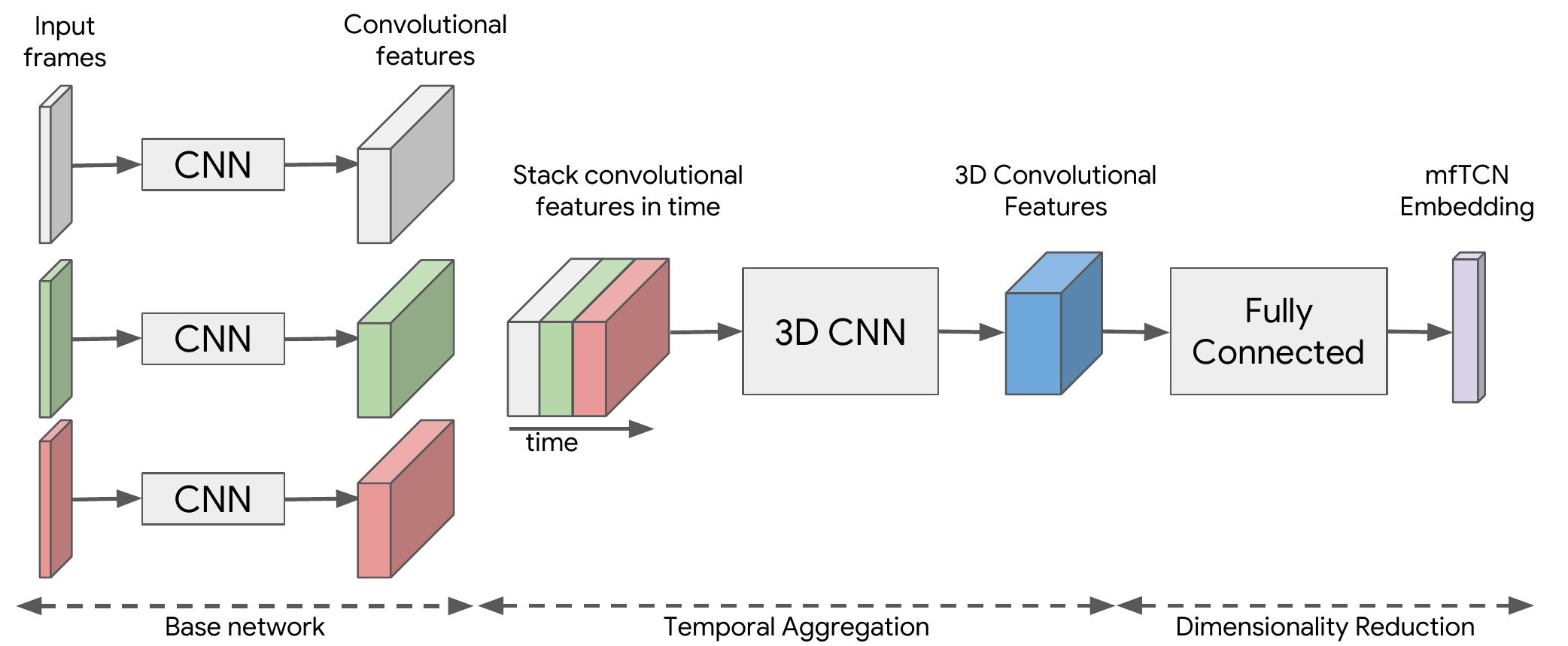}
  \caption{Architecture used to extract mfTCN embeddings from a given sequence of frames. The 3D convolution layer is added to aggregate temporal information across frames and encode motion cues in the mfTCN embedding.}
  \label{fig:architecture}
\end{figure*}

We present the architecture of the deep network used in mfTCN in Figure~\ref{fig:architecture}. This network takes a clip that is a sequence of frames as input and jointly embeds them in a low-dimensional space. We describe different parts of the network below:

\subsubsection{Base Network}
We use a convolutional neural network (CNN) as the base network which will be used to extract low-dimensional representations from raw pixels. We will learn mfTCN embeddings on top of these representations. 
The base network encodes the hidden features $h_t$ at time t from an input frame $I_t$:
$$
h_t = CNN(I_t)
$$

\subsubsection{Temporal Aggregation}
There are several approaches that can be used to aggregate temporal information from features extracted from frames at different time steps. We can perform temporal averaging, use temporal convolutions~\cite{tran2015learning,carreira2017quo} or use recurrent neural networks~\cite{donahue2015long}. We choose to use 3D temporal convolutions to perform temporal aggregation in our model but other architectures may also be used.

$$
\phi_t = Conv3D(h_t,h_{t-s}...,h_{t-(n-1) \times s})
$$

\subsubsection{Dimensionality Reduction}
After the temporal aggregation step, we end up with a convolutional feature map that has 4 dimensions(time, height, width, channels). In order to learn a compact representations of entire clips, we reduce the dimensionality by first perfoming spatial averaging and then using a fully connected layer.

$$
mfTCN_t = FC(\phi_t)
$$

\subsubsection{Time-Contrastive Learning}
We use the n-pairs loss introduced in \cite{sohn2016improved} to train our network. For explanation purposes, we borrow the terms ``anchor, positive and negative" from the triplet loss~\cite{Schroff2015Unified} literature because similar concepts are used for the n-pairs loss. In general, an anchor-positive pair is a pair of samples that we want to be closer to each other in embedding space than the anchor-negative pair.  The loss aims at learning a metric embedding which clusters data samples of the same category closer to each other while pushing them farther from samples from different category in the learned embedding space. To be able to use the n-pairs loss in a time-contrastive setting, we sample non-overlapping clips at the same timestep from multiple views. We obtain $k$ examples (equal to the number of views) at each timestep. These examples are considered as positives. If we have sampled at $n$ timesteps, then we get $k\times (n-1)$ examples are negatives for the $k$ positives at each timestep (See Figure~\ref{fig:sampling}). In other words, clips sampled from different views but at the same time are positive while clips sampled at a different time are considered as negative irrespective of the view. This heuristic provides the required supervisory signal to train our model. In Figure~\ref{fig:sampling}, we can see that with just two samples from two views we are able to generate 6 supervisory labels: 2 pairs of embeddings that should be close to each other and 4 pairs of embeddings that should be distant from each other. This number grows combinatorially with the number of timesteps we sample in a given batch. This rich supervisory signal coupled with the context provided by embedding multiple frames jointly allows us to learn good representations from visual observations.

\subsubsection{Architectural Differences between TCN and mfTCN}
\label{sec:tcn_differences}
There are a couple of architectural differences between the original TCN~\cite{sermanet2017time} and mfTCN. First, there is a 3D convolution layer in mfTCN that takes convolutional feature maps from different time steps as input. Second, we do not have a spatial softmax~\cite{finn2015learning} layer in mfTCN which was used by the original TCN to reduce dimensionality.

\subsection{Learning Control Policies}

In order to test the utility and robustness of our embeddings we decide to perform continuous control tasks on top of these learned representations. Continuous control algorithms usually take the true state (joint angle, positions etc) as input. We want our learned representation to be a drop-in replacement for the true states required for a particular task. In order to do so , we choose PPO~\cite{schulman2017proximal} as the on-policy optimization algorithm that allows us to learn continuous control policies.

\section{Experiments}
\label{sec:experiments}

\subsection{Regression to velocities and positions of Cartpole}
We aim to show that the mfTCN embeddings are able to encode velocities and positions of objects and parts of agents. It is important to measure how well our model is able to capture this information because if we expect to be able to use these embeddings to perform continuous control tasks they need to be encoding precise information about these entities. 

In order to measure this quantitatively, we collect multi-view videos of the \textit{Cartpole} environment from the Deepmind Control Suite~\cite{tassa2018deepmind}. We use the default multiple camera setup in their environment. We learn a multi-frame TCN embedding from the videos collected where the agent is taking random actions. The random actions follow the same distribution as \cite{jonschkowski2017pves}. We evaluate if these embeddings are able to encode the position and velocity of different objects present in the scene. In order to do so, we follow the experimental setup of ~\cite{jonschkowski2017pves} in which they train regressors on top of their embeddings using the true states provided by the simulator. Like Position Velocity Encoders, we also concatenate the difference between embeddings at time $t$ and time $(t-1)$ along with our embedding. We train 3 fully connected layers of 256 dimensions on top of the learned embedding with an Adam optimizer and learning rate of 0.001. The trained regressors are then shown frames from the validation set and predict the true states of the agent. We use a train/test split of 160000 and 20000 frames for this task. The results of this experiment are reported in Table~\ref{table_regression}. We observe that the multi-frame embeddings are able to encode the true position and velocity better. Additionally, the Mean Squared Error (MSE) decreases when we increase the size of the embedding learned. Interestingly, we do not explicitly train for the embedding to encode position or velocity but it stands to reason the model needs to encode them, among other attributes, to differentiate between clips at different timesteps while still knowing that multiple views at the same time should encode the same state.  

\begin{table*}[h]
\caption{Results on regressing Cartpole Positions and Velocities from mfTCN Embeddings}
\label{table_regression}
\begin{center}
\begin{tabular}{c c || c c c|c c c| c c }
 &  &   \multicolumn{3}{c|}{MSE}   &  \multicolumn{3}{c|}{MSE in Static Attributes} &	 \multicolumn{2}{c}{MSE in Motion Attributes}\\
\hline
Embedding & Lookback  & Average  & Position  & Motion  &		&  &	 &   & \\
size  & window  &  &   &   &       $x_{cart}$ &	$sin (\theta_{pole})$	& $cos (\theta_{pole})$ & $\dot{x}_{cart}$ &	$\dot{\theta}_{pole}$\\
\hline
\hline
8 & 1 & 0.1312 & 0.0201 & 0.2979 & 0.0238 & 0.0085 & 0.0279 & 0.4964 & 0.0993\\
8 & 2 & 0.0435 & 0.0017 & 0.1062 & 0.0011 & 0.0025 & 0.0016 & 0.1587 & 0.0537\\
8 & 4 & \textbf{0.0207} & \textbf{0.0014} & \textbf{0.0497} & 0.0008 & 0.0020 & 0.0014 & 0.0788 & 0.0207\\
\hline
32 & 1 & 0.0911 & 0.0052 & 0.2201 & 0.0045 & 0.0037 & 0.0072 & 0.3716 & 0.0686\\
32 & 2 & 0.0401 & 0.0019 & 0.0974 & 0.0014 & 0.0027 & 0.0015 & 0.1456 & 0.0492\\
32 & 4 & \textbf{0.0198} & \textbf{0.0013} & \textbf{0.0476} & 0.0008 & 0.0020 & 0.0013 & 0.0748 & 0.0203\\
\hline
128 & 1 & 0.0636 & 0.0040 & 0.1528 & 0.0030 & 0.0035 & 0.0056 & 0.2396 & 0.0661\\
128 & 2 & 0.0419 & 0.0018 & 0.1020 & 0.0011 & 0.0027 & 0.0015 & 0.1489 & 0.0552\\
128 & 4 & \textbf{0.0211} & \textbf{0.0014} & \textbf{0.0505} & 0.0007 & 0.0022 & 0.0014 & 0.0783 & 0.0227\\
\hline

\end{tabular}
\end{center}
\end{table*}

\subsection{Policy learned on TCN embedding from Self-Observation}
In this experiment we consider the scenario where an agent is able to observe itself in its environment. We consider the \textit{Swing-up} task in the Cartpole domain of Deepmind's Control Suite. The task is to swing the pole up by applying forces at the base of the cart and then to balance it without deviating too much from the center of the base. The physical model is similar to the one presented in ~\cite{barto1983neuronlike}. In order to collect data to train the mfTCN, the agent performs random actions with random initial states. We use the default camera setup. Note in this setup the second camera is moving with the agent.

We use the Proximal Policy Optimization (PPO)~\cite{schulman2017proximal} algorithm to learn a policy on top of the learned mfTCN embeddings. Usually the true states (like positions of objects, velocities of joints etc.) are provided as input to learn a policy. We replace the true states with the learned low-dimensional representation  as input to PPO. We report the mean and standard deviation of rewards for 100 episodes/rollouts at test time. Following \cite{tassa2018deepmind}, we rollout for 1000 steps for each episode.

The results of this experiment are reported in Table~\ref{table_cartpole}. We observe that if we train both the representation learning network and the policy network in parallel on raw pixels the agent is not able to learn the task(row 3). As a baseline, we use embeddings from the Position Velocity encoders to train a PPO policy and that results in the agent performing better(row 4). We observe that the mfTCN trained on one frame at the same resolution(row 5) is able to outperform both raw pixels and PVE as input by a large margin. The performance improves significantly when we jointly embed 5 frames(row 6). This highlights the advantages of embedding multiple frames jointly as opposed to only using a single frame. We find performance of mfTCN improves again when we provide input at a higher resolution(rows 7 and 8). 

Since, we learned an MFTCN model from multiple views, we do not need to retrain another representation learning model to train another policy for the second camera, which in this case is moving. We use the same mfTCN network and train another PPO policy and find that the version of mfTCN which embeds 5 frames jointly(row 10) works much better with the moving camera as compared to the one which embeds only one frame(row 9).

\begin{table}[h]
\caption{Results on CartPole Swingup task}
\label{table_cartpole}
\begin{center}
\begin{tabular}{l |c c c | c c }
             & \# look- &       &  & \multicolumn{2}{c}{Cumulative}\\
             & back & From  &  & \multicolumn{2}{c}{Reward}\\
Input to PPO & Frames & Pixels  & Resolution & Mean & Std.\\
\hline
\hline
Random State &  1& & & 121.45 & 11.98\\
True State & 1 && & 861.41 & 3.46\\
\hline
Pixels (CNN) &  1 & \checkmark& $96\times48$ & 283.82 &	42.88\\
PVE~\cite{jonschkowski2017pves} & 1& \checkmark& $96\times48$ & 457.27 & 51.16\\
mfTCN &  1&\checkmark &$96\times48$ & 550.98 & 53.40	\\
mfTCN & 5&\checkmark & $96\times48$ & 701.30 & 80.14	\\
\hline
mfTCN & 1&\checkmark & $160\times160$ & 759.33	& 77.77  \\
mfTCN & 5&\checkmark & $160\times160$ & 787.47 & 67.80\\

\hline 
mfTCN (moving) & 1& \checkmark & $160\times160$ & 691.77& 85.28\\
mfTCN (moving) & 5 &\checkmark & $160\times160$ & 811.10 &	41.80\\

\end{tabular}

\end{center}
\end{table}

\subsection{Policy learned on TCN embedding from Observing Other Agents}

Contrary to the previous experiment, we consider the scenario where an agent is able to observe other similar agents performing a given task. In particular, we consider the \textit{Cheetah} environment which is a tougher control task than \textit{Cart-pole} in terms of having a larger state space and possible number of actions. We are provided with demonstration videos of the \textit{Cheetah} agent walking successfully. These demonstrations were generated by training a PPO policy on the true state of the agent. We use the default camera setup for this environment. We train our mfTCN model on these videos only. One manner in which these demonstrations differ from the previously considered scenario  where an agents performs random actions is that the agent only ever sees successful examples of walking and does not see the plethora of states a Cheetah encounters while it is learning to walk. This is similar to the real-life imitation learning settings where it is easy to gather successful demonstrations. The big question is: can we learn a representation useful for learning the task at hand by only observing successful demonstrations? With this experiment we show that it is possible to do so for the \textit{Walk} task in the $Cheetah$ domain. Similar to the above section, we report the mean and standard deviation of rewards for 100 episodes/rollouts at test time. Following \cite{tassa2018deepmind}, we rollout for 1000 steps for each episode.

\begin{table}[h]
\caption{Results on Cheetah Walk task}
\label{table_example}
\begin{center}
\begin{tabular}{l | c c }
             & \multicolumn{2}{c}{Cumulative}\\
             & \multicolumn{2}{c}{Reward}\\
Input to PPO & Mean & Std.\\
\hline
\hline
Random State &   28.31 & 3.62\\
True State & 390.16 & 44.85	\\
\hline
Pixels (CNN) &  146.14 & 29.51\\
mfTCN & 360.50 &  76.52\\

\end{tabular}
\end{center}
\end{table}

\begin{table}[h]
\caption{Attributes classification and alignment metrics on the Pouring Dataset: mfTCN consistently outperforms other baselines.}
\label{table:mftcn_pouring}
\begin{center}
\begin{tabular}{lc||c c c}
 & Lookback  &  Static  & Motion & Alignment\\
Model& window  &  Error (\%) & Error (\%)  & Error (\%)  \\
\hline
\hline
Random & 1 & 54.2 & - & -\\
Inception & 1 & 51.9 & - & -\\
Shuffle\&Learn\cite{misra2016shuffle} & 1 &  27.0 & - & -\\
TCN~\cite{sermanet2017time} & 1 & 22.2 & -& - \\
\hline
mfTCN$^1$ & 1 &  18.9 & 30.2 & 16.2\\
mfTCN$^7$ & 7 &  17.3 & 24.8 & 14.3\\
mfTCN$^{13}$ & 13 &  \textbf{16.8} & \textbf{18.3} & 11.3 \\
mfTCN$^{29}$ & 29  & 19.4 & 20.9 & \textbf{8.9}\\
\end{tabular}
\end{center}
\end{table}

\begin{table*}[h]
\caption{Attribute-wise classification results on the Pouring Dataset}
\label{table:mftcn_pouring_details}
\begin{center}
\begin{tabular}{lc|c c c c c | c c c c}
  &     &  \multicolumn{5}{c|}{Error in Static Attributes (\%)} &	 \multicolumn{4}{c}{Error in Motion Attributes (\%)}\\
\hline
 & Lookback  &  & &  &	has &	liquid &	container &	container &	hand &	hand\\
Model & window  &    contact &	angle	& distance &	liquid &	flowing &	down &up &	reaching &	receding\\
\hline
\hline

Random & 1 & 49.9 &  74.5  & 48.9 & 49.2 & 48.4 & - & - & - & - \\
Inception & 1 & 47.4 &   71.8  &  45.2 &48.8 & 49.2 & - & - & - & - \\
Shuffle \& Learn~\cite{misra2016shuffle} & 1 & 17.2 &  46.3 & 17.8 & 25.7 & 28.0 & - & - & - & - \\
TCN~\cite{sermanet2017time} & 1 & 8.0 &  \textbf{35.9} & \textbf{9.0} & 24.7 & 35.5 & - & - & - & - \\
\hline
mfTCN$^{1}$ & 1  & \textbf{7.2} & 39.5 & 9.5 & 27.4 & 14.6 & 34.5 & 30.6 & 30.5 & 29.6 \\
mfTCN$^{7}$ & 7 &  8.2 & 39.7 & 10.6 & 22.1 & 10.3 & 25.7 & 25.6 & 25.8 & 25.0 \\
mfTCN$^{13}$ & 13 & 8.2 & 38.7 & 9.1 & \textbf{18.5} & \textbf{10.2} & \textbf{22.6} & \textbf{19.7} & \textbf{17.2} & \textbf{17.6} \\
mfTCN$^{29}$ & 29 & 8.6 & 37.5 & 11.2 & 16.1 & 10.3 & 25.4 & 20.3 & 16.2 & 20.7\\
\hline
\end{tabular}
\end{center}
\end{table*}
\subsection{Classification results on Pouring dataset}
To evaluate the usefulness of our learned representations on real-world data we conduct experiments on the \textit{Pouring} dataset introduced in~\cite{sermanet2017time}. Additionally this allows us to compare the mfTCN model with its TCN cousin using the results reported in~\cite{sermanet2017time}. The dataset consists of videos of humans pouring different liquids into cups. The dataset has two synchronized views of the captured scene: a fixed first-person view and a moving third person view (see \fig{sampling} for an example). We use a pre-trained InceptionV3~\cite{szegedy2016rethinking} as the base network and extract convolutional features from \textit{Mixed\_5d} to train the mfTCN model (we do not fine-tune the weights below that as in~\cite{sermanet2017time}). In particular, we are interested in observing the effect of varying the size of lookback window of mfTCN on the classification of attributes relevant for the \textit{Pouring} task. we describe those attributes in detail below.

In the original dataset, frames are manually labeled with answers to questions that should be pertinent to the task of pouring. The original dataset had the following questions and answers: \begin{enumerate}
  \item Is the hand in contact with the container? (yes or no)
  \item Is the container within pouring distance of the recipient? (yes or no)
  \item What is the tilt angle of the pouring container? (90, 45, 0 and -45 degrees)
  \item Is the liquid flowing? (yes or no)
\item Does the recipient contain liquid? (yes or no)
\end{enumerate}

These questions, though relevant to the task, are restricted to information contained only in a single-frame. We augment the set of questions with motion based questions that are also relevant to pouring but will typically require more than 1 frame to answer. The motion based questions are all binary questions (yes or no):
\begin{enumerate}
  \item Is the hand reaching towards the container?
  \item Is the hand receding from the container?
  \item Is the container in contact with the hand going up?
  \item Is the container in contact with the hand coming down?
\end{enumerate}

Note that these labels are only used for the purposes of evaluation but never during training. Following the experimental setup of~\cite{sermanet2017time}, we use a nearest neighbor classifier to quantitatively evaluate the usefulness of the learned embeddings. Given a sequence from the evaluation set, we retrieve the embedding nearest neighbor of each frame in the set of embeddings from all frames of all other sequences of the evaluation set (the current sequence is excluded). The labels of the selected nearest neighbor are used to classify the attributes for the current frame.
We also report the an alignment error metric that measures how well two views of a video are aligned in time. We take advantage of the fact that frames in both videos are in sync with each other to get alignment labels for free. For any two views of the same demonstration, we first embed all frames using mfTCN. Then for each frame in the first view we look for the nearest neighbours in the second view.  We retrieve the time index of the nearest neighbours ($time_{knn}$) to see how well they are aligned. The alignment error for a particular frame $i$ is defined as follows:

$$
error_i = \frac{abs(|time_i - time_{knn}|)}{sequence\ length}
$$

In \tabl{mftcn_pouring}, we report the attributes classification and average alignment error metrics on the Pouring dataset. We report classification error for static and motion attributes separately to identify better the contributions of the mfTCN model over the TCN baseline (which operates with only 1 frame as input). We also report in more details the error rates of each attribute in \tabl{mftcn_pouring_details}.
We find that mfTCN outperforms other baselines and obtains the best classification results with a window of size $13$ (mfTCN$^{13}$). We report the results found in~\cite{sermanet2017time} and compare mfTCN$^{1}$ to the TCN baseline. These two models are equivalent in that they both use a single frame as input, however their architectures differ (details in \sect{tcn_differences}) which explains the observed performance difference. As expected, a window size greater than $1$ yields significant improvements in classification of motion attributes (from $30.2\%$ error down to $18.3\%$ error with mfTCN$^{13}$). More surprisingly perhaps we also observe an error reduction for the static attributes. One explanation can be found in \tabl{mftcn_pouring_details} by observing that attributes "has liquid" and "liquid flowing" get a significant boost of performance, we hypothesize that these attributes can be recognized both using static and motion cues, hence taking motion into account helps. For the alignment error, we find that the model with the largest input window (mfTCN$^{29}$) performs best. One interpretation is that it is easier to disambiguate clips in the embedding space if there is more context available to the model. Overall, we find that mfTCN consistently outperforms the TCN baseline.

\subsection{Discussion}
We show that our approach can encode the proprioceptive states of an agent from pixels. Like ~\cite{sermanet2017time}, we expect the approach to also learn rich representations about relevant objects in the environment. One drawback of the present approach is that the embedding can choose to fixate on some objects in the environment while ignoring others. Even though in our experiments we only used the learned embeddings to learn control policies, in a more practical setting one should use both the embedding as well as the proprioceptive states as input. Although our representation learning approach is self-supervised, it still relies on being presented with a reasonable coverage of possible states. Such issues may be alleviated with an  explicit exploration strategy (like intrinsic motivation~\cite{barto2013intrinsic}) or expert demonstrations. 

\section{Conclusion}

In this paper, we extended TCN by allowing it to embed multiple frames jointly. We show that by doing so we get better estimates of positions and velocities which leads to better performance in continuous control tasks. Embedding multiple frames also leads to significant performance improvement on the \textit{Pouring} dataset.

We show that this approach to learning robust visual representations allows us to use policy learning algorithms effectively directly from video as opposed to true states. The results on the simulated environments are encouraging and we aim to use this model to learn more robust policies on real robots. In the future, we also want to be able to refine the representation learned based on any new data that the agent encounters after taking actions.

\section*{Acknowledgement}
We thank Sergey Levine and Vincent Vanhoucke for reviews and constructive feedback. We are grateful to Martin Riedmiller  and  Rico  Jonschkowski  for  their  help  with  the Position-Velocity Encoder models.

\bibliographystyle{IEEEtran}
\bibliography{IEEEabrv,mybibfile}

\end{document}